\documentclass{SVProc}
\usepackage[utf8]{inputenc}
\usepackage{pgfplots}
\DeclareUnicodeCharacter{2212}{−}
\usepgfplotslibrary{groupplots,dateplot}
\usetikzlibrary{patterns,shapes.arrows}
\usetikzlibrary{positioning}
\pgfplotsset{compat=newest}
\usepackage{makeidx}  % allows for indexgeneration
\usepackage{acro}
\usepackage{amsmath}
\usepackage{amssymb}
\usepackage{booktabs}
\usepackage{fontawesome}
\usepackage{graphicx}
\usepackage{multirow}
\usepackage{calc}
\usepackage{hyperref}
\usepackage{tikz}
\usepackage{microtype}
\usepackage{caption}
\usepackage{subcaption}
\usepackage[doi=false,url=false,isbn=false,style=ieee,date=year]{biblatex}
\usepackage{enumitem}
\setlist{nosep}

\newcommand\blfootnote[1]{%
  \begingroup
  \renewcommand\thefootnote{}\footnote{#1}%
  \addtocounter{footnote}{-1}%
  \endgroup
}

\makeindex

\newbibmacro{string+doiurlisbn}[1]{%
  \iffieldundef{doi}{%
    \iffieldundef{url}{%
      \iffieldundef{isbn}{%
        \iffieldundef{issn}{%
          #1%
        }{%
          \href{http://books.google.com/books?vid=ISSN\thefield{issn}}{#1}%
        }%
      }{%
        \href{http://books.google.com/books?vid=ISBN\thefield{isbn}}{#1}%
      }%
    }{%
      \href{\thefield{url}}{#1}%
    }%
  }{%
    \href{http://dx.doi.org/\thefield{doi}}{#1}%
  }%
}

\DeclareFieldFormat{title}{\usebibmacro{string+doiurlisbn}{\mkbibemph{#1}}}
\DeclareFieldFormat[article,incollection,inproceedings]{title}%
    {\usebibmacro{string+doiurlisbn}{\mkbibquote{#1}}}
\DeclareFieldFormat[article,incollection,inproceedings]{publisher}%
    {}
\DeclareSourcemap{
  \maps[datatype=bibtex]{
    \map{
       \perdatasource{references.bib}
   \pertype{article}
   \pertype{book}
   \pertype{unpublished}
   \pertype{inproceedings}
      \step[fieldset=publisher, null]
      \step[fieldset=pages, null]
      \step[fieldset=note, null]
    }
  }
}

\usepackage{url}

\DeclareMathOperator*{\miou}{mIoU}

\newcommand*\circled[1]{\tikz[baseline=(char.base)]{
            \node[shape=circle,draw,inner sep=1pt] (char) {#1};}}

\DeclareAcronym{slam}{
 short = SLAM ,
 long = simultaneous localisation and mapping,
}
\DeclareAcronym{iou}{
 short = IoU ,
 long = intersection over union,
}
\DeclareAcronym{crf}{
 short = CRF ,
 long = conditional random field,
}

\begin{document}
%\bstctlcite{IEEEexample:BSTcontrol}
%
%\frontmatter          % for the preliminaries
%
\pagestyle{headings}  % switches on printing of running heads
%%%\addtocmark{Hamiltonian Mechanics} % additional mark in the TOC
%
%
\title{SCIM: Simultaneous Clustering, Inference, and Mapping for Open-World Semantic Scene Understanding}
\titlerunning{Simultaneous Clustering, Inference, and Mapping}  % abbreviated title (for running head)
%                                     also used for the TOC unless
%                                     \toctitle is used
%
\author{Hermann Blum\inst{1}, Marcus G Müller\inst{1,2}, Abel Gawel\inst{3}, Roland Siegwart\inst{1},\\ and Cesar Cadena\inst{1}}
\authorrunning{Hermann Blum et al.} % abbreviated author list (for running head)
%
%%%% list of authors for the TOC (use if author list has to be modified)
%\tocauthor{Ivar Ekeland, Roger Temam, Jeffrey Dean, David Grove, Craig Chambers, Kim B. Bruce, and Elisa Bertino}
%
%\index{Ekeland, I.}
%\index{Temam, R.}
%\index{Dean, J.}
%\index{Grove, D.}
%\index{Chambers, C.}
%\index{Kim, B.}
%\index{Bertino, E.}
%
\institute{Autonomous Systems Lab, ETH Zürich, Switzerland
\and
German Aerospace Center (DLR), Munich, Germany
\and
Huawei Research, Zürich, Switzerland\blfootnote{This paper was financially supported by the HILTI Group.}
}

\maketitle              % typeset the title of the contribution

\begin{abstract}
%The abstract should summarize the contents of the paper using at least 70 and at most 150 words. It will be set in 9-point font size and be inset 1.0 cm from the right and left margins. There will be two blank lines before and after the Abstract. 
In order to operate in human environments, a robot's semantic perception has to overcome open-world challenges such as novel objects and domain gaps. 
Autonomous deployment to such environments therefore requires robots to update their knowledge and learn without supervision.
We investigate how a robot can autonomously discover novel semantic classes and improve accuracy on known classes when exploring an unknown environment. To this end, we develop a general framework for mapping and clustering that we then use to generate a self-supervised learning signal to update a semantic segmentation model. In particular, we show how clustering parameters can be optimized during deployment and that fusion of multiple observation modalities improves novel object discovery compared to prior work. Models, data, and implementations can be found at \href{https://github.com/hermannsblum/scim}{\url{github.com/hermannsblum/scim}}.
\end{abstract}

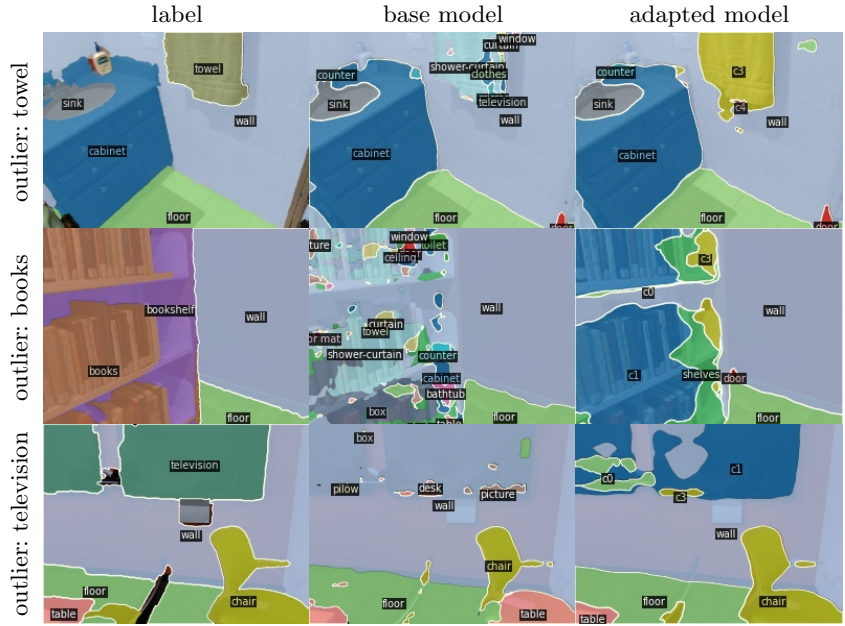
\begin{figure}[t]
    \centering
\vspace{-2mm}
\begin{tikzpicture}[
every node/.style={outer sep=0},
node distance = 0mm,
label distance = -2mm,
]
    \node[label={above:label\strut}] (firstrow) at (0, 0) {\includegraphics[width=100pt]{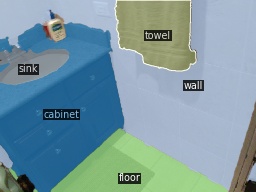}};
\node[left=of firstrow]{\rotatebox{90}{outlier: towel}};
    \node[label={base model\strut}] at (3.5, 0) {\includegraphics[width=100pt]{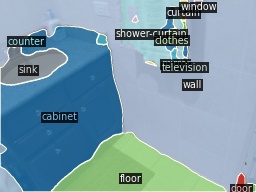}};
    \node[label={adapted model\strut}] at (7, 0) {\includegraphics[width=100pt]{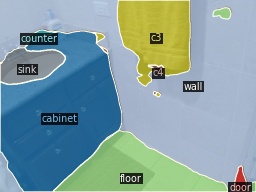}};
    
    \node (secondrow) at (0, -2.6) {\includegraphics[width=100pt]{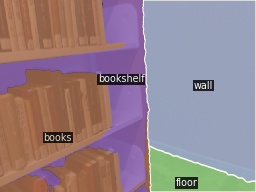}};
\node[left=of secondrow]{\rotatebox{90}{outlier: books}};
    \node at (3.5, -2.6) {\includegraphics[width=100pt]{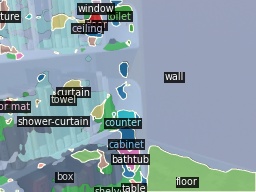}};
    \node at (7, -2.6) {\includegraphics[width=100pt]{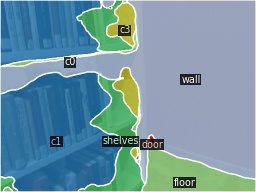}};
    
    \node (thirdrow) at (0, -5.2) {\includegraphics[width=100pt]{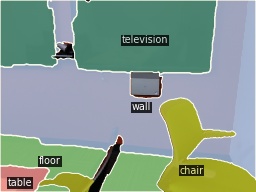}};
\node[left=of thirdrow]{\rotatebox{90}{outlier: television}};
    \node at (3.5, -5.2) {\includegraphics[width=100pt]{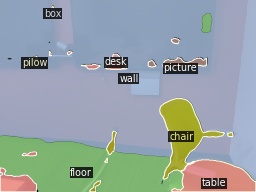}};
    \node at (7, -5.2) {\includegraphics[width=100pt]{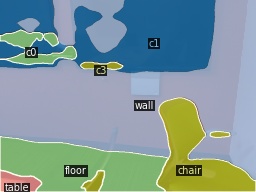}};
\end{tikzpicture}
\vspace{-10pt}
    \caption{Example predictions from a segmentation network trained with our method (third column), compared to the predictions of a pretrained model (second column) and the ground-truth label (first column). The base model did not see the outlier class during training. These novel objects are discovered autonomously, and are therefore not labelled by a word but by their cluster id (e.g., \texttt{c1}, \texttt{c2}, ...).}
    \label{fig:examples}
\vspace{-5mm}
\end{figure}
\section{Introduction}
Robots that automate tasks such as household work, hospital logistics, assistive care, or construction work have to operate in environments that are primarily designed for humans. Moreover, all these tasks have a high level of complexity that requires semantic scene understanding~\cite{garg_semantics_2020}. Semantics in human environments are open-world. They have domain gaps, contain novel objects, and change over time. For robots to operate autonomously in such environments, they need to be able to deal with such changes and novelties. This requires a methodological shift from deploying models trained on fixed datasets to enabling robots to learn by themselves, building up on advancements in self-supervised learning and continual learning. It is the robotic version of ideas like `learning on the job'~\cite{liu_learning_2020}, open-world object detection~\cite{joseph_towards_2021} or segmentation~\cite{cen_deep_2021}, and is related to the idea of developmental robotics~\cite{lungarella_developmental_2003}.

This work investigates scenarios where robots should perform semantic scene understanding in unknown environments that contain novel object categories. We show how robots can improve their semantic segmentation in these new environments on both known and unknown categories by leveraging semantic mapping, uncertainty estimation, self-supervision, and clustering. We call the investigated problem `Simultaneous Clustering, Inference, and Mapping' (SCIM). The approaches we investigate work fully autonomously without human supervision or intervention. While fusion of predictions and discovery of novel objects has also been investigated in the context of semantic mapping~\cite{mccormac_semanticfusion_2016,grinvald_volumetric_2019}, maps are always bound to one point in time and one specific environment. Instead, segmentation networks can carry knowledge into different environments. Prior work has shown that combining self-supervised pseudo-labels with continual learning can integrate knowledge gathered over multiple environments in closed-world~\cite{blum_self-improving_2021}, which would not be possible with mapping alone. Therefore, we investigate promising self-supervision signals for open-world class-incremental learning.

Based on the motivation to deploy robots to human environments, we focus this work on indoor scenes. Given a trajectory in the unknown environment, the robot collects observations with a RGB-D sensor in different modalities. These encompass (1) predictions of the base segmentation model and their uncertainties, (2) deep features of the segmentation network and other image-based networks, (3) a volumetric map of the environment, and (4) geometric features extracted from that map. Subsequently, we categorize these observations through clustering and integrate everything back into the semantic segmentation network by training it with pseudo-labels. We obtain an updated network that can identify novel semantic categories and has overall higher prediction accuracy in the environment, as shown in Figure~\ref{fig:examples}. 
In summary, our contributions are:
\begin{itemize}
    \item A novel framework to map, discover, represent, and integrate novel objects into a semantic segmentation network in a self-supervised manner.
    %\item A framework for information collection in unknown environments that generalises over several existing approaches for semantic mapping and novel object discovery.
    \item An algorithm that leverages prior knowledge to optimise the clustering parameters for finding representations of novel objects.
    \item We provide the first open-source available method implementations and, while not at the  scale of a benchmark, develop metrics and evaluation scenarios for open-world semantic scene understanding.
    %\item An experimental setup to test novel object discovery and scene adaptation in indoor environments.
    %\item a new best-overall method 
\end{itemize}
\section{Related Work}
\label{sec:related}
%\subsection{Novel Object Discovery}
\subsubsection{Novel object discovery} describes an algorithm's ability to account for unknown object classes in perception data. Grinvald et al.~\cite{grinvald_volumetric_2019} showed a semantic mapping framework that was able to segment parts of the scene as `unknown object', but without the ability to categorize these. Nakajima et al.~\cite{nakajima_incremental_2019} were one of the first to demonstrate semantic scene understanding that can identify novel objects. They rely on superpixel segmentation, mapping, and clustering to identify object categories.  Hamilton et al.\cite{hamilton_unsupervised_2022} showed fully unsupervised video segmentation based on a similar framework like the one we describe in Section~\ref{sec:problem_theory}. Both~\cite{nakajima_incremental_2019} and~\cite{hamilton_unsupervised_2022} cluster a whole scene into semantic parts without the ability to relate a subset of clusters to known labels. Uhlemeyer et al.~\cite{uhlemeyer_towards_2022} recently demonstrated that based on advancement in out-of-distribution detection, a segmentation can be split into inliers and outliers. They cluster only the outliers into novel categories, but based on features that were already supervised on some of their outlier classes. In contrast to our method, theirs is therefore not fully unsupervised and further lacks the capability to integrate multiple observation modalities.

%\subsection{Clustering}
\subsubsection{Clustering}\label{sub:rel:clustering} for classification can be understood as a two part problem. First, (high dimensional) descriptors for the items in question have to be found. Then, a clustering algorithm groups similar descriptors together. The established approach in representation learning is to learn a single good descriptor that can be clustered with kNN or k-means~\cite{caron_emerging_2021}.
K-means can be used with mini-batches, is differentiable, fast, and easy to implement. However, we argue that there are two big disadvantages: it requires a priori knowledge of the number of clusters $k$ and only works in the space of a single descriptor. 
For a scenario in which a robot is deployed to unknown environments, it cannot know the number of novel object categories. It is also questionable whether a single descriptor will be able to well describe all parts of the unknown environment.
Graph clustering algorithms like DBSCAN~\cite{ester_density-based_1996} or HDBSCAN~\cite{campello_density-based_2013} can cluster arbitrary graphs and do not need to know the number of clusters beforehand. Their hyperparameters are instead related to the values of the connectivity matrix. These graph edges are independent of any specific descriptor space and can e.g. also be an average over multiple descriptor distances. This makes graph clustering a better candidate when working with more than one descriptor (in clustering literature, this is called `multi-view' clustering~\cite{fu_overview_2020}, which however is a very ambiguous term in scene understanding). Unfortunately, there exist no differentiable graph clustering algorithms. The closest in the literature is~\cite{shah_deep_2018}, which however links nodes to a single descriptor. \cite{du_differentiable_2021}
proposes a differentiable multi-descriptor clustering, which however just uses representation learning to link from multiple to one descriptor. Our work therefore investigates how graph clustering can be used to cluster scenes based on multiple descriptors. We further investigate how parameters of graph clustering can be tuned without gradient based optimisation.

\subsubsection{Continual learning} describes the problem of training a single model over a stream of data that e.g. contains increasing amount of classes, shifts in data distribution, etc. New knowledge should be integrated into the model without forgetting old knowledge. Prior work~\cite{blum_self-improving_2021,uhlemeyer_towards_2022} has already shown the effectiveness of pseudo-labels in continual learning and other works~\cite{yu_self-training_2020,michieli_incremental_2019} showed techniques for supervised class-incremental semantic segmentation that mitigate forgetting. However, supervision is not available in autonomous open-world operation. We therefore focus this work on the self-supervision part of class-incremental learning, while referring to~\cite{yu_self-training_2020,michieli_incremental_2019} for ways to address or evaluate forgetting.

\section{Method}
In the following we define the problem of open-world scene understanding. We first take a step back and set up a general formal description of the problem, which we show relates to similar works on segmentation, and enables us to identify the core differences between the evaluated methods. We then describe how we identify novel object categories, and how we use this identification to train the robot's segmentation network in a self-supervised manner.

\begin{figure}[t]
    \centering
    \vspace{-1mm}
    \includegraphics[width=.6\linewidth]{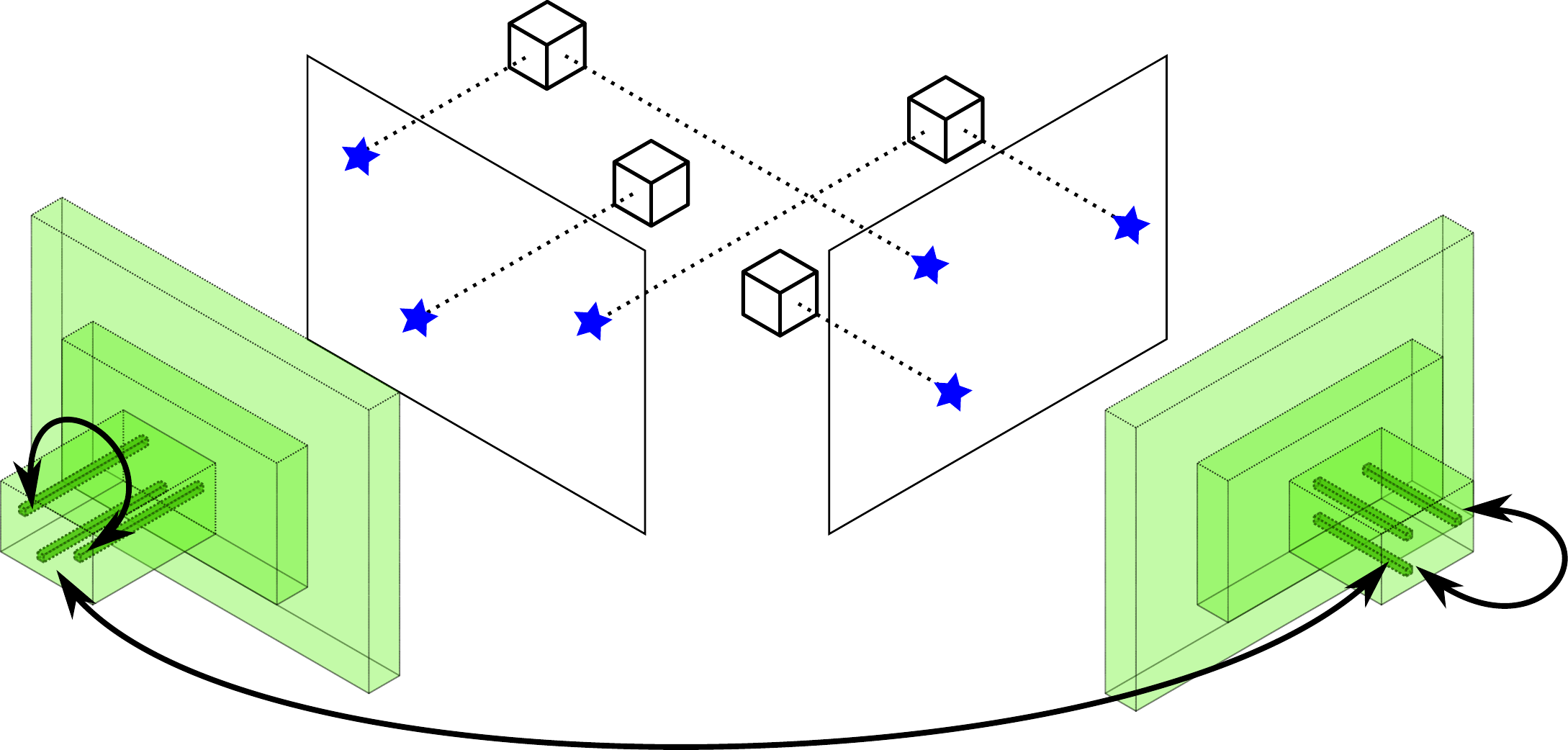}
    \vspace{-1mm}
    \caption{Illustration of the graph setup. Every node (star) is a pixel in a camera frame. For every node, we also know the projection into 3D and potential correspondences from other frames. We can further relate nodes within and between frames through deep features from networks run on the frames.}
    \label{fig:graph_illustration}
    \vspace{-4mm}
\end{figure}

\subsection{Preliminaries: Scene Understanding as a Clustering Problem}
\label{sec:problem_theory}

A robot explores an environment and collects over its trajectory camera images and consequently through \ac{slam} corresponding poses in relation to the built map. 

Let $G = (V, E)$ describe a graph where every vertex $v_i \in V$ is an observed pixel, as illustrated in Fig.~\ref{fig:graph_illustration}. For each of these observation vertices $v_i$, the following information is available:
\begin{itemize}
  \setlength\itemsep{0pt}
    \item image plane coordinates in the corresponding camera image
    \item 3D coordinates $T_{\textrm{map} \rightarrow v_i}$ in the robot map
    \item time of observation $t(v_i)$
    \item semantic prediction $\textrm{pred}(v_i)$ and associated (un)certainty $\textrm{cert}(v_i)$
    \item any local visual / learned / geometric feature $f(v_i)$ that can be inferred from the corresponding camera image, map location, or additional sensing modalities available on the robot
\end{itemize}
Edges can then in general be found from a function $e: V \times V \rightarrow \mathbb{R}_{\geq 0}$:
\begin{align*}
    e(v_i, v_j) = e(T_{v_i \rightarrow v_j},\; \langle f(v_i), f(v_j)\rangle,\;
    \langle t(v_i), t(v_j) \rangle,\; \langle\textrm{pred}(v_i), \textrm{pred}(v_j)\rangle)
\end{align*}
that, based on distance functions $\langle\cdot, \cdot\rangle$, distills the multimodal relations of $v_i$ and $v_j$ into an edge weight. 

The semantic interpretation of the scene can then be expressed as the graph clustering problem on $G$ that splits into disjoint clusters $V_k$. 
\begin{align*}
V = \bigcup \limits_{\textrm{K clusters}} V_k \qquad \forall i, j, i \neq j: V_i \cap V_j = \O
\end{align*}

The formulation above is very related to \acp{crf} and the unsupervised segmentation loss from Hamilton et al.~\cite{hamilton_unsupervised_2022}.  They show that this graph structure represents a Potts problem~\cite{potts_generalized_1952}. Briefly summarised: If $\phi: V \rightarrow \mathcal{C}$ (softly) assigns vertices to clusters and $\mu: \mathcal{C} \times \mathcal{C} \rightarrow \mathbb{R}$ sets a cost for the comparison between two assignments, e.g. the cross-entropy, the graph clustering problem introduced above minimizes the following energy term:
\begin{align*}
    E(\phi) = \sum \limits_{v_i, v_j \in V} e(v_i, v_j) \mu (\phi(v_i), \phi(v_j))
\end{align*}

In the most simple case of perfect (i.e. ground-truth) predictions,
 \begin{align*}
     e(v_i, v_j) = \left\lbrace \begin{array}{ll}
         \textrm{const.} &\textrm{if pred}(v_i) = \textrm{pred}(v_j)\\
         0 & \textrm{otherwise}
    \end{array}\right.
 \end{align*} is a disjoint graph with each cluster matching one label.
 
Often however, predictions are not perfect and techniques like volumetric semantic mapping can be used to filter noisy predictions by enforcing that a voxel should have the same class regardless of the viewpoint. This roughly corresponds to the problem of clustering $G$ with e.g. 
\begin{align*}
    e(v_i, v_j) = \left\lbrace \begin{array}{ll}
         1 & \textrm{if }v_i, v_j \textrm{ in same voxel}\\
         <1 & \textrm{if pred}(v_i) = \textrm{pred}(v_j)\\
         0 & \textrm{otherwise}
    \end{array}\right.
\end{align*}
where some approaches also take uncertainty or geometry into account.

\subsection{Identifying Novel Categories}
\label{sub:clustering}
To identify novel categories in a scene, observation vertices $v_i$ need to be clustered based on commonalities that go beyond position in the map and prediction of a pretrained classifier, because by definition the pretrained classifier will not be able to identify novel categories. As the predominant current approach in category prediction is to identify categories visually, we also follow this approach to cluster observations into novel categories based on a range of visual descriptors. In the above introduced graph clustering framework, this means:
\begin{align}
    e_\textrm{ours}(v_i, v_j) = \sum \limits_{d \in \textrm{descriptors}} w_d \langle f_d(v_i), f_d(v_j)\rangle \label{eq:edge_ours}
\end{align}
with $w_d$ the weight for each descriptor and $\sum w_d = 1$. Note that equation~(\ref{eq:edge_ours}) is a generalisation of different related work. For example, Uhlemeyer et al.\cite{uhlemeyer_towards_2022} links observations only by the feature of an Imagenet pretrained ResNet and Nakajima et al.~\cite{nakajima_incremental_2019} weight features from the pretrained segmentation network and geometric features based on the entropy of the classification prediction $h(v_i)$.
\begin{align*}
    e_\textrm{uhlemeyer}(v_i, v_j) =& \; \|\textrm{tSNE}(\textrm{PCA}(f_\textrm{imgnet}(v_i))) - \textrm{tSNE}(\textrm{PCA}(f_\textrm{imgnet}(v_j)))\|_2  \\
    e_\textrm{nakajima}(v_i, v_j) =& \; \|(1-h(v_i)) f_\textrm{segm}(v_i) - (1-h(v_j)) f_\textrm{segm}(v_j)\|_2 \\&+ \|h(v_i) f_\textrm{geo}(v_i) - h(v_j) f_\textrm{geo}(v_j)\|_2
\end{align*}
where $\textrm{tSNE}(\textrm{PCA}(\cdot))$ is a dimensionality reduction as described in~\cite{uhlemeyer_towards_2022}.

\subsubsection{Optimisation of Clustering Parameters}\ \\
To solve the clustering problem of $G(V, E)$, different hyper parameters $\Theta$ have to be found. This includes parameters of the clustering algorithm and the weights $w_d$ of the different descriptors. The choice of these parameters governs the `grade of similarity' that is expected within a cluster, i.e. how fine-grained categories should be. In general, choosing these parameters is very hard because the choice has to hold for unknown objects and scenes. In this work, we therefore propose to solve the parameter choice by optimisation. In particular, we propose to use the subset $\tilde{V} \in V$ of observations where the prediction $\textrm{pred}(v)$ out of one of the known classes $1, ..., K$ has high certainty $\forall v \in \tilde{V}: \textrm{cert}(v) > \delta$ and find the hyperparameters $\Theta$ as follows:
\begin{align}
    \Theta = \textrm{argmax} \; \; \miou \limits_{v \in \tilde{V}} \left( \lbrace V_1, ..., V_K \rbrace, \lbrace \textrm{pred}(v) = 1, ..., \textrm{pred}(v) = K \rbrace \right) \label{eq:param-optimisation}
\end{align}
where mIoU is the mean \ac{iou} of the best matching between the clustering $\lbrace V_1, ..., V_K \rbrace$ and the predictions of the pretrained classifier. %Because we measure this objective only over observations  with low uncertainty, we trust these predictions to be close to the truth.

As shown in Section~\ref{sec:related}, there exists no graph clustering algorithm that is differentiable either to its input or its parameters. Eq.~(\ref{eq:param-optimisation}) is therefore not optimisable based on gradients. Hence, we employ black-box optimisation that models the clustering algorithm as a gaussian process $\Theta \rightarrow \miou$. Based on the `skopt' library~\cite{head_tim_scikit-optimizescikit-optimize_2021}, samples of $\Theta$ are choosen to cover the optimiation space but favoring areas where good mIoU is expected based on previous measurements. After 200 iterations, we select the point with the best mIoU.

\subsubsection{Subsampling of the Clustering Graph}\ \\
Usually $G(V, E)$ is too large to efficiently compute the clustering problem. With an already low image resolution of 640x480 and a frame rate of e.g. 20Hz, 2 min of camera trajectory correspond to $|V| > 7e9$. We therefore rely on random subsampling, taking 100 random points on every 5th frame, to create a smaller problem that is still representing the whole scene. Such subsampling removes redundancies in observations of neighboring pixels and subsequent frames, but also exaggerates noise that would otherwise average out over more data points. We hence choose parameters to the maximum possible with the available memory.

Subsampling however comes with the challenge that there is no direct clustering solution for all $v \in V$. We therefore combine the subsampled graph clustering with nearest neighbor search. Given a clustering for a subsampled part of $V$, we assign every $v$ either to its cluster, if it was part of the graph clustering, or to the cluster of the nearest neighbor according to $e_\textrm{ours}(v_i, v_j)$.

Note that prior work reduces computationally complexity by first segmenting the scene into superpixels or segments and then clustering these. This approach comes with other challenges, notable how to attribute features to segments and how to ensure segments are not merging independent objects. With our subsampling, we test an alternative approach.

\subsubsection{Self-Supervised Class-Incremental Training}\ \\
\label{sub:selfsupervision}%
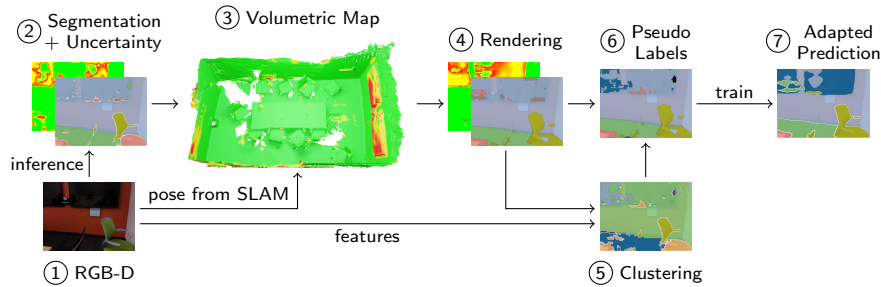
\begin{figure}[t]%
\vspace{-2mm}
    \centering
\begin{tikzpicture}[
every node/.style={outer sep=1pt, inner sep=1pt},
node distance = 3.5mm,
font=\scriptsize\sffamily,
]
\node[label={below:\circled{1} RGB-D}] (rgbd) {\includegraphics[width=1.2cm]{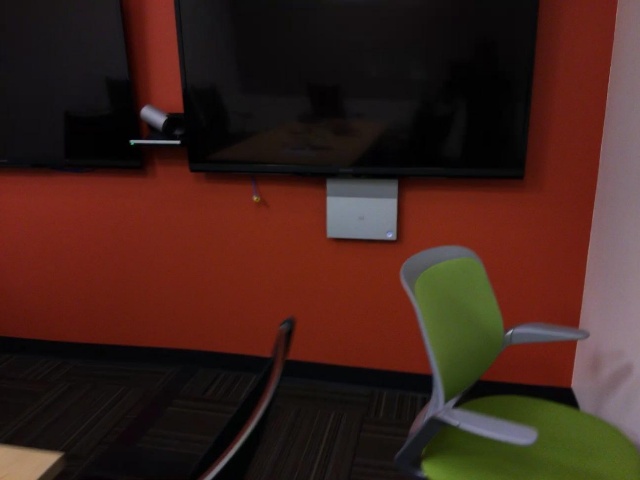}};

\node[above=of rgbd, label={\circled{2} \parbox{\widthof{Segmentation}}{Segmentation + Uncertainty}}] (pred) {
\begin{tikzpicture}[every node/.style={outer sep=0pt, inner sep=0pt}]
\node at (-1.5mm, 1.5mm) {\includegraphics[width=1.2cm]{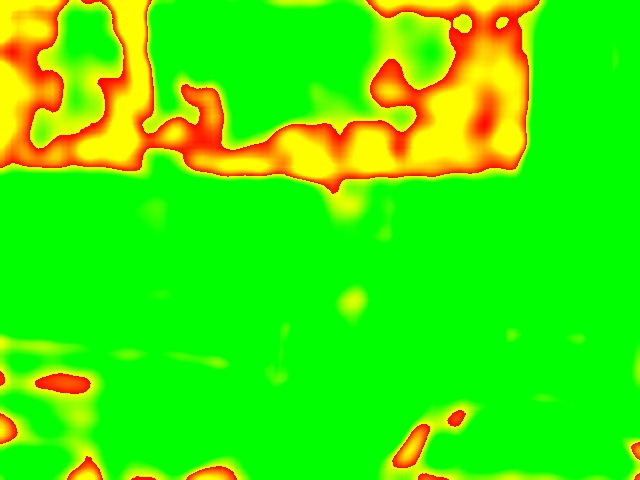}};
\node at (1.5mm, -0.5mm) {\includegraphics[width=1.2cm]{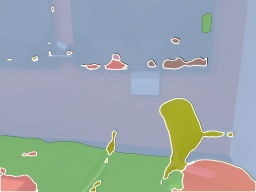}};
\end{tikzpicture}
};

\node[label={\circled{3} Volumetric Map}, right=of pred] (map) {\includegraphics[width=3cm]{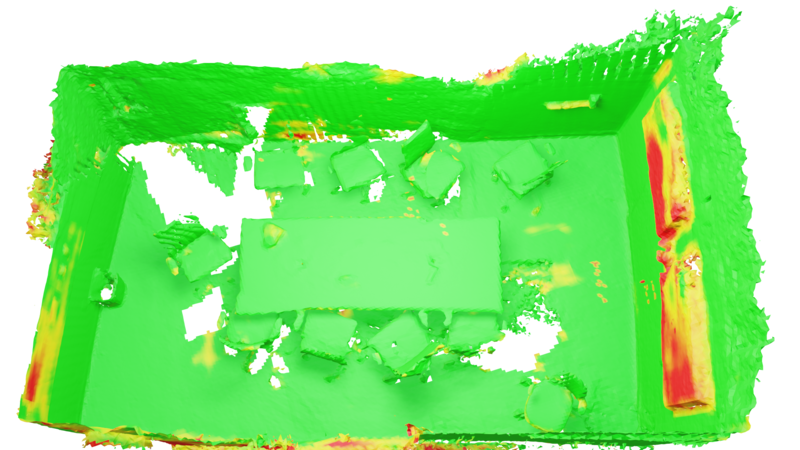}};

\node[label={\circled{4} Rendering}, right=of map] (render) {
\begin{tikzpicture}[every node/.style={outer sep=0pt, inner sep=0pt}]
\node at (-1.5mm, 1.5mm) {\includegraphics[width=1.2cm]{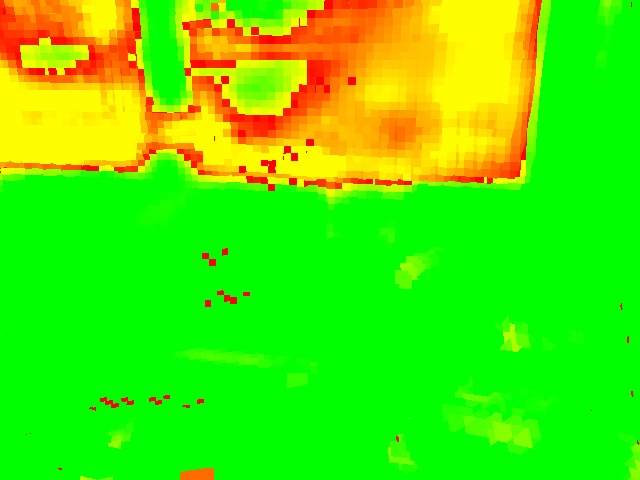}};
\node at (1.5mm, -0.5mm) {\includegraphics[width=1.2cm]{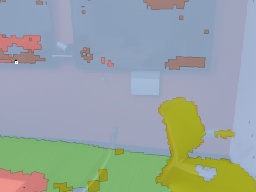}};
\end{tikzpicture}
};

\node[label={\circled{6} \parbox{\widthof{Pseudo}}{\centering Pseudo Labels}}, right=of render] (label) {\includegraphics[width=1.2cm]{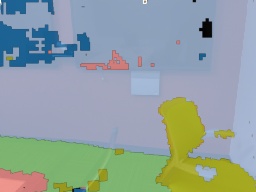}};

\node[label={below:\circled{5} Clustering}, below=4.5mm of label] (cluster) {\includegraphics[width=1.2cm]{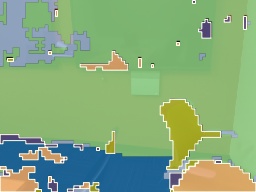}};

\node[label={\circled{7} \parbox{\widthof{Prediction}}{\centering Adapted Prediction}}, right=1cm of label] (adapted) {\includegraphics[width=1.2cm]{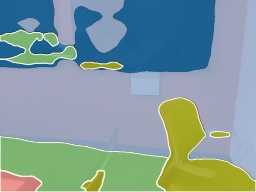}};

\draw [->] (rgbd)->(pred) node[midway,left] {inference};
\draw [->] (pred)->(map);
\draw [->] ([yshift=1mm]rgbd.east)-|(map) node[pos=.25,above] {pose from SLAM};
\draw [->] (map)->(render);
\draw [->] (render.south)|-([yshift=1mm]cluster.west);
\draw [->] ([yshift=-1mm]rgbd.east)--([yshift=-1mm]cluster.west) node[pos=.5,below] {features};
\draw [->] (render.east)->(label.west);
\draw [->] (cluster)->(label);
\draw [->] (label)->(adapted) node[midway,above] {train};
\end{tikzpicture}
\vspace{-3mm}
    \caption{Overview of the steps for self-supervised, class-incremental learning. The method is described in Section~\ref{sub:selfsupervision} and implementation details in Section~\ref{sub:methodimplementation}.}
    \label{fig:systemdiagram}
    \vspace{-5mm}
\end{figure}%
To adapt the robot's perception to a novel scene, on both known and unknown categories, we leverage the above described method as part of a larger system that is outlined in Figure~\ref{fig:systemdiagram}. Its goal is to produce a useful learning signal without any supervision, i.e. purely based on the observations the robot makes itself~\circled{1}. With such a learning signal, we finetune the base segmentation model~\circled{2} with the goal of improving the predictions in the given scene on the next trajectory.
%Overall, we follow these steps:
%\begin{enumerate}
%    \item inference of semantic segmentation and associated uncertainty
%    \item pose estimation and volumetric semantic mapping of the scene
%    \item rendering of the mapped semantics and uncertainties back into the %camera frames
%    \item clustering as described in Section~\ref{sub:clustering}
%    \item compilation of pseudo-labels
%    \item finetuning of the base model with the pseudo-labels
%\end{enumerate}

As a volumetric mapping framework~\circled{3}, we use the implementation from~\cite{schmid_panoptic_2022} without the panoptic part. This framework performs volumetric mapping and integration of semantic predictions per voxel. We extend this framework to also integrate uncertainties associated with the predictions as average per voxel. For pose-estimation, we rely on the poses provided with the data, which are obtained through visual-inertial mapping and bundle adjustment~\cite{dai_scannet_2017}.

Because the semantic map integrates many predictions from different viewpoints, it cancels out some noise from single frame predictions and has a higher overall accuracy. This was used in~\cite{frey_continual_2021} as a learning signal for scene adaptation, but assuming that all classes are known. Similar to~\cite{frey_continual_2021}, we also render~\circled{4} the semantic map back into each camera pose, obtaining an improved semantic prediction for every frame. We additionally render the averaged uncertainty, which indicates which parts of the scene are reliable predictions on inlier classes (low average uncertainty) and which parts are either novel categories or unfamiliar known categories (both high uncertainty), as shown in Figure~\ref{fig:uncertainty_maps}.

Given a clustering solution~\circled{5}, we produce pseudo-labels for training by merging renderings from the semantic map with clustering-based predictions~\circled{6}. We first identify clusters with large overlap to predicted categories by measuring the contingency table between the semantic classes in the map and the clustering. We merge clusters that have an \ac{iou} $>.5$ with the corresponding predicted class. All other clusters are considered novel categories. We then assign for each pixel in each camera frame, i.e. each $v_i$, either (i) the rendered semantic class from the map if the average map uncertainty is below a threshold $\delta$
or (ii) the assigned cluster if the uncertainty of the prediction is higher than $\delta$.
Essentially, our pseudolabels therefore identify unknown parts of a scene and produce a learning signal to perform domain adaptation for the known parts of the scene and novel object discovery for the unknown parts of the scene.

To train the model on the pseudo labels~\circled{7}, we need to extend its last layer to accomodate the newly identified categories. We do this by increasing the kernel and bias of the last layer from $\mathbb{R}^{ F \times C}$ to $\mathbb{R}^{F \times (C + C')}$ where $F$ is the feature size, $C$ is the number of known classes, and $C'$ is the number of novel detected clusters. We then initialise these matrixes with standard random initialisation for the new rows and with the base model's parameters for the already existing rows.

\section{Experimental Evaluation}
%In the following we describe how we evaluate novel object discovery and scene adaptation for indoor robotic applications. In particular, we explain our choice of dataset, networks, and pretraining. We then discuss suitable metrics for evaluation and finally present experimental results.

%\subsection{Problem Setup}
We investigate the setting where a robot is put into a new environment that contains objects it has never seen. We choose the ScanNet dataset~\cite{dai_scannet_2017} of RGB-D trajectories in real indoor environments. Despite multiple errors in the semantic annotations, ScanNet had the highest data quality when we searched for datasets containing RGB-D trajectories, poses, and semantic annotations. The dataset is split into scenes (usually one room), where each scene may include multiple trajectories.
%We then select three outlier classes that models must not be trained on. These classes are chosen according to their scarcity in the ScanNet dataset, which would otherwise remove large parts of the training data (wall, table, door, etc. appear in many scenes). The classes also need to appear often enough such that there are enough validation scenes to test the methods.
We select television, books, and towel as outlier classes, which the segmentation models must not have seen during training. After data quality filtering to e.g. account for incorrect annotations, we end up with 3 scenes with TVs, 1 scene with books and 2 scenes with towels (out of the first 100 validation scenes). Compared to prior work that tests on a total  of 360 images~\cite{nakajima_incremental_2019} and 951 images~\cite{uhlemeyer_towards_2022}, we therefore test on significantly more data with total of 15508 frames and at least 1000 frames per trajectory.

\subsection{Method Implementation}
\label{sub:methodimplementation}
We run two variants of our method: One is only taking  self-supervised information as input, i.e. features of the segmentation network itself, self-supervised visual features from DINO~\cite{caron_emerging_2021}, and geometric features. The second variant (following~\cite{uhlemeyer_towards_2022}) is also fusing visual information of a ResNet101 trained on Imagenet as input. Imagenet features are obtained by supervised training on a wide range of classes, including the ones we consider as outliers in our experimental setting, so they cannot be considered self-supervised.

We obtain geometric features by running the provided model of~\cite{gojcic_perfect_2019}, which trains a descriptors to register 3D point clouds, on the surface point cloud of the voxel map. We extract features of our segmentation network at the `classifier.2' layer, which is one layer before the logits. As Imagenet features, we take the output of the last ResNet block before flattening, such that the features still have spatial information (`layer4' in pytorch). From DINO, we take the last token layer that still has spatial relations. We normalise these descriptors to a l2 norm of 1 and calculate pairwise euclidean distances. We then harmonize the scale of different feature distances by finding scalar factors $\alpha$ such that $p(\alpha * |v_i - v_j| < 1) = .9$ for $v_i, v_j$ that have low uncertainty and the same predicted class.

For clustering, we use HDBScan~\cite{campello_density-based_2013}. This is an improved version of DBScan that is designed to deal better with changing densities of the data and we found it in general to perform slightly more reliable.% than DBScan when using edge weights that fuse many different feature distances. 

As segmentation network, we use a DeepLabv3+ trained on COCO and then on ScanNet, but not on the scenes or objects categories we test on. We employ standard image augmentation and random crops. For uncertainty estimation, we take the method with the best simplicity-performance trade off from the Fishyscapes benchmark~\cite{blum_fishyscapes_2021}: We use the max-logit value of the softmax, including the post-procesing introduced in~\cite{jung_standardized_2021}, but without their standardisation step.%. We do not use their standardisation step, because it did not scale well to 40 classes.% When training on pseudo-labels, we train for 30 epochs with a reduced learning rate of $0.00001$. Further implementation details can be found in our code and the appendix.

\subsection{Adaptation of Baselines}
In addition to the full optimisation based approach above, we want to test concepts from related work. Unfortunately, neither~\cite{nakajima_incremental_2019} nor~\cite{uhlemeyer_towards_2022} made their implementation available, which is why we implement both methods and adapt them to our evaluation setting. It is important to note that these are not direct replications or reproductions, but we take as many ideas from these papers  as possible and combine them with the system above to get the best result.

For Nakajima et al.~\cite{nakajima_incremental_2019}, we implement their clustering procedure, but take the same segmentation network, mapping framework, and geometric features as for our method. Since we could not find all clustering parameters in their paper, we run our proposed parameter optimisation on the inflation and $\eta$ parameter of the MCL clustering for every scene. Because MCL takes longer than HDBScan, we needed to use stronger subsampling. Given that our geometric features are very different than the ones used in~\cite{nakajima_incremental_2019}, we also evaluate a variant that only uses the features of the segmentation network. In~\cite{nakajima_incremental_2019}, this had slightly worse performance.

For Uhlemeyer et al.~\cite{uhlemeyer_towards_2022}, their underlying meta-segmentation is released as open-source. However, their uncertainty estimation is only available for an urban driving network, so we replace it with the uncertainty metric that we also use for our method. Also this paper does not report its clustering parameters. We however cannot run our proposed parameter optimisation for this method, because it only clusters the outliers. We therefore, following advice of the authors, hand-tune the parameters until we get a good result for scene 0354 and use these settings ($\epsilon = 3.5$, min samples $= 10$) for all scenes.

\subsection{Evaluation Metrics}
The evaluation protocol in prior work is not well documented~\cite{nakajima_incremental_2019} or limited to one novel cluster~\cite{uhlemeyer_towards_2022}. We therefore describe our protocol in more detail.
Class-incremental learning is usually supervised and therefore evaluated with a standard confusion matrix~\cite{douillard_plop_2020}. In unsupervised representation learning literature, the number of clusters is usually set to the number of labels, such that the Hungarian Algorithm can find the optimal matching of clusters and labels~\cite{caron_emerging_2021,hamilton_unsupervised_2022}.

In our problem setting, these assumptions do not hold. Parts of the predicted classes are trained in a supervised manner and only the novel categories are found in an unsupervised way. Because parts of the scenes contain outlier objects without annotation, we can also not punish a method for finding more categories than there are labels. To match clusters to existing labels, we first count how often which cluster is predicted for which label in the contingency matrix of size $N_\textrm{labels}\times N_\textrm{clusters}$ on the labelled part of the scene. We then use a variant of the Hungarian algorithm~\cite{noauthor_munkres_nodate} that first pads the matrix with zeros to a square matrix and further disallows to assign predictions with a supervised label to another label (they may however be disregarded entirely if an unlabelled cluster is a better match). If no prediction has a supervised label, this is equivalent to the established Hungarian matching from representation learning. If all predictions have supervised labels, this is equivalent to the standard confusion matrix.

We also report the v-score~\cite{rosenberg_v-measure_2007}, which is independent of label-cluster matching. It is the harmonic mean over two objectives: All points in a cluster belong to the same label, and all predictions of a label come from the same cluster.

\subsection{Results}
    We present the main results of this study in Table~\ref{tab:results}. Qualitative examples can be found in the video attachment. In general, we observe that all methods are able to detect the novel object categories to a certain degree, except for variants of nakajima in scene 0598. We also observe that the full nakajima variant with segmentation and geometric features performs poorly, but note that the method was originally designed for different geometric features and the segmentation-only variant performs much better. %Given that also our optimisation-based approach assigns very low weight to the geometric features in most scenes, this indicates that the used geometric features are not well informative of semantics. %This could either be because they are too local (only describing a specific edge of an object) or too global (describing an entire part of the scene with multiple objects).

\begin{table}
    \centering
\resizebox*{!}{.8\textheight}{%
    \begin{tabular}{lllrrrr}
\toprule
 & & & \multicolumn{2}{c}{training traj.} & \multicolumn{2}{c}{new traj.}\\ 
outlier & scene & method & out IoU & mIoU & out IoU & mIoU\\
\midrule
\multirow{7}{*}{tv} & \multirow{7}{*}{0354} & 
base model & 0 & 47 & - & -\\
& & adapted nakajima: segm. + geom. & 7 & 4 & - & -\\
& & adapted nakajima: segm. only & 9 & 20 & - & -\\ % 3440
%& & ours unsupervised & \faCheck & 1 & - &  & 0 & \textbf{60} & \textbf{58} & - & - \\ % 3478
& & SCIM fusing segm. + geom. + dino & \textbf{73} & \textbf{62} & - & - \\ % 3524
& & adapted uhlemeyer: imgn. & 65 & 60 & - & -\\ % 3457 
 &  & adapted uhlemeyer: imgn. + map & 21 & 56 & - & -\\ % 3211
& & SCIM fusing segm. + geom. + imgn. & 41 & 54 & - & -\\ % 3403
\midrule
\multirow{7}{*}{tv} & \multirow{7}{*}{0575} & 
base model & 0 & 53 & 0 & 53\\
& & adapted nakajima: segm. + geom. & 10 & 11 & - & -\\ % 3351
& & adapted nakajima: segm. only & 24 & 17 & - & -\\ % 3062
% & & ours unsupervised & \faCheck & .98 & - &  & .02 & 0 & \textbf{60} & 0 & \textbf{57}\\ % 3068
 & & SCIM fusing segm. + geom. + dino & \textbf{33} & \textbf{64} & \textbf{39} & \textbf{63}\\ % 3526
 &  & adapted uhlemeyer: imgn. & 12 & 55 & 23 & 61\\ % 3096
 &  & adapted uhlemeyer:  imgn. + map & 6 & 62 & 4 & 58\\ % 3353
 & & SCIM fusing segm. + geom. + imgn. & \textbf{50} & \textbf{69} & 28 & \textbf{63}\\ % 3060
\midrule
\multirow{7}{*}{tv} & \multirow{7}{*}{0599} & 
base model & 0 & 60 & 0 & 63\\
& & adapted nakajima: segm. + geom. & 1 & 0 & - & -\\
& & adapted nakajima: segm. only & 36 & 26 & - & -\\
% & & ours unsupervised & \faCheck & 1 & - &  & 0 & \textbf{47} & \textbf{69} & \textbf{45} & \textbf{65}\\ % 3477
& & SCIM fusing segm. + geom. + dino & \textbf{37} & \textbf{67} & \textbf{36} & \textbf{63}\\ % 3491
 &  & adapted uhlemeyer: imgn. & 42 & \textbf{71} & \textbf{55} & \textbf{74}\\ % 3230
 &  & adapted uhlemeyer: imgn. + map & 11 & 70 & 10 & 68 \\ % 3227
& & SCIM fusing segm. + geom. + imgn. & 32 & 65 & 31 & 62\\ % 3405
\midrule
\multirow{7}{*}{books} & \multirow{7}{*}{0598} & 
base model & 0 & 59 & - & -\\
& & adapted nakajima: segm. + geom. & 1 & 1 & - & -\\ % 3349
& & adapted nakajima: segm. only & 0 & 0 & - & -\\ % 3475
% & & ours unsupervised & \faCheck & .94 & - &  & .06 & \textbf{34} & \textbf{74} & - & - \\ % 3272
& & SCIM fusing segm. + geom. + dino & \textbf{43} & \textbf{77} & - & - \\ % 3529
 &  & adapted uhlemeyer: imgn. & 29 & 63 & - & -\\ % 3279
 &  & adapted uhlemeyer: imgn. + map & 29 & 70 & - & -\\ % 3275
 &  & SCIM fusing segm. + geom. + imgn. & 33 & 74 & - & -\\ % 3274
\midrule
\multirow{7}{*}{towel} & \multirow{7}{*}{0458} & 
base model & 0 & 48 & 0 & 38\\
& & adapted nakajima: segm. + geom. & 6 & 7 & - & -\\ % 3469
& & adapted nakajima: segm. only & \textbf{63} & 44 & - & -\\ % 3468
%& & ours unsupervised & \faCheck & .98 & - &  & .02 & \textbf{70} & \textbf{61} & \textbf{77} & \textbf{51}\\ % 3293
& & SCIM fusing segm. + geom. + dino & 33 & \textbf{57} & \textbf{37} & \textbf{47}\\ % 3547
& & adapted uhlemeyer: imgn. & 38 & 61 & 47 & 49 \\ % 3281
& & adapted uhlemeyer: imgn. + map & 38 & 59 & 46 & 48 \\ % 3365
& & SCIM fusing segm. + geom. + imgn. & 60 & \textbf{63} & \textbf{79} & \textbf{55}\\ % 3298
\midrule
\multirow{7}{*}{towel} & \multirow{7}{*}{0574} & 
base model & 0 & 45 & - & -\\
& & adapted nakajima: segm. + geom. & 22 & 7 & - & -\\ % 3352
& & adapted nakajima: segm. only & 4 & 27 & - & -\\ % 3488
%& & ours unsupervised & \faCheck & .94 & - & - & .06 & \textbf{29} & \textbf{56} & - & -\\ %3467
& & SCIM fusing segm. + geom. + dino & \textbf{28} & \textbf{52} & - & -\\ %3527
& & adapted uhlemeyer: imgn. & 23 & 50 & - & - \\ % 3321
& & adapted uhlemeyer: imgn. + map & f & f & - & - \\ % 
& & SCIM fusing segm. + geom. + imgn. & \textbf{39} & \textbf{53} & - & -\\ % 3335
\bottomrule
\end{tabular}}\vspace{3pt}
    \caption{Model predictions in \lbrack \% mIoU\rbrack\ for different scenes and different outlier classes. We mark the best unsupervised method and the best overall. The available unsupervised information are the base model's features (segm.), geometric features (geom.), and DINO~\cite{caron_emerging_2021} features (dino). Some methods however use features from supervised ImageNet training (imgn.). Note that `nakajima' is a pure clustering method, so we measure the clustering instead of model predictions. For those scenes where a second trajectory is available, we evaluate trained models on the second trajectory.}
    \label{tab:results}
\end{table}

\begin{figure}[t]
    \centering
\begin{tikzpicture}[
every node/.style={outer sep=0, inner sep=0},
node distance = 0mm,
]
\node (1) {\includegraphics[height=2cm, trim={0cm 0 0cm 0}, clip]{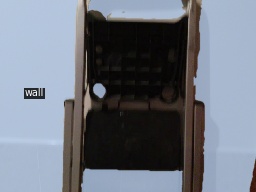}};
\node[right=of 1] (2) {\includegraphics[height=2cm, trim={0cm 0 0cm 0}, clip]{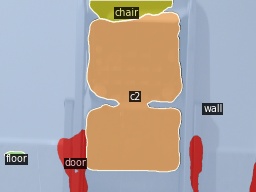}};

\node[right=1cm of 2] (3) {\includegraphics[height=2cm, trim={0cm 0 0cm 0}, clip]{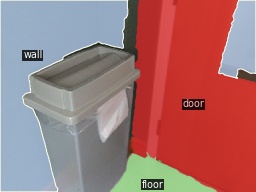}};
\node[right=of 3] {\includegraphics[height=2cm, trim={0cm 0 0cm 0}, clip]{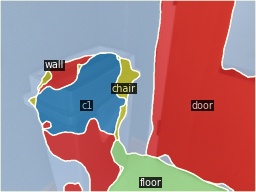}};
\end{tikzpicture}
\vspace{-3mm}
    \caption{Predictions on unlabelled parts of the scene reveal that the adapted network detects more novel classes than the labels allow to measure.}
    \label{fig:more_objects}
    \vspace{-4mm}
\end{figure}

Those methods that adapt the segmentation model through training (ours and uhlemeyer) in all cases improve performance over the base model. Where a second trajectory of the environment is available, we can also verify that this is a true improvement of the segmentation and not overfitting on the frames. Notably, our fully unsupervised variant\footnote{`unsupervised' refers to novel classes. All tested methods are supervised on the known classes.} is better than all supervised methods in 2 scenes and competitive in the other scenes. This indicates that the use of supervised ImageNet pretraining is limited and very useful features can instead be learned unsupervisedly from any environment in open-world deployment.

As discussed in Section~\ref{sub:rel:clustering}, graph clustering does not require a priori knowledge of the number of classes. As such, Figure~\ref{fig:more_objects} shows examples of objects that are not measureable as outliers, yet got discovered by the algorithms and predicted as a new category by the trained network.

In scenes like 0568 or 0164, no investigated approach is able to discover the outlier class. Especially small objects and cluttered scenes pose big challenges.

We conclude that our approach that is fusing multiple sources of information, especially the unsupervised variant, shows the most consistent performance over the listed scenes, but there is no approach that is the best in every scene.

\begin{figure}[t]
    \centering
\begin{subfigure}[t]{0.25\textwidth}
\vspace{0mm}
    \centering
    \includegraphics[height=2cm, trim={7cm 2cm 5cm 4cm}, clip]{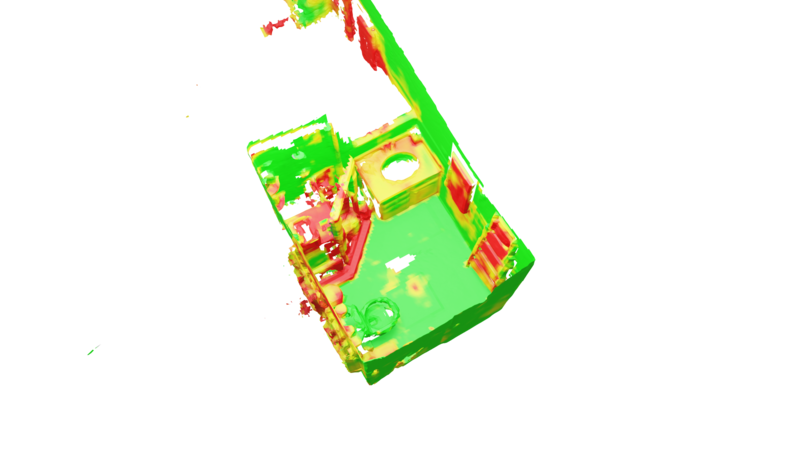}
    \caption{Scene 0458}
\end{subfigure}
\begin{subfigure}[t]{0.25\textwidth}
\vspace{0mm}
    \centering
    \includegraphics[height=2cm, trim={5cm 0cm 5cm 2cm}, clip]{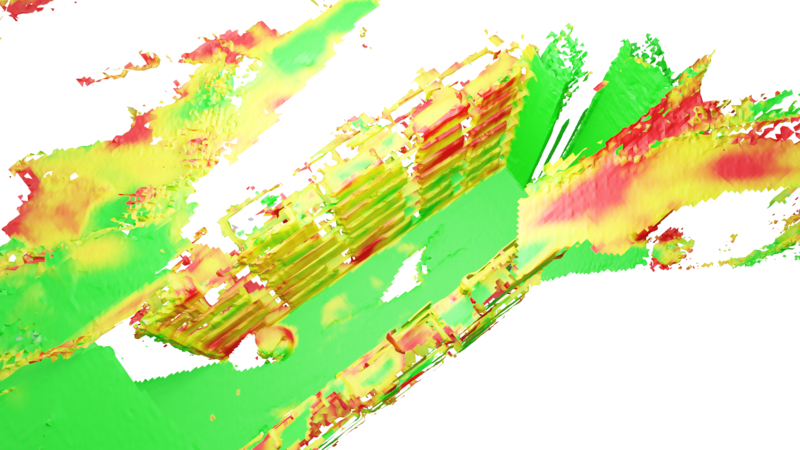}
    \caption{Scene 0598}
\end{subfigure}
\begin{subfigure}[t]{0.3\textwidth}
\vspace{0mm}
    \centering
    \includegraphics[height=2cm]{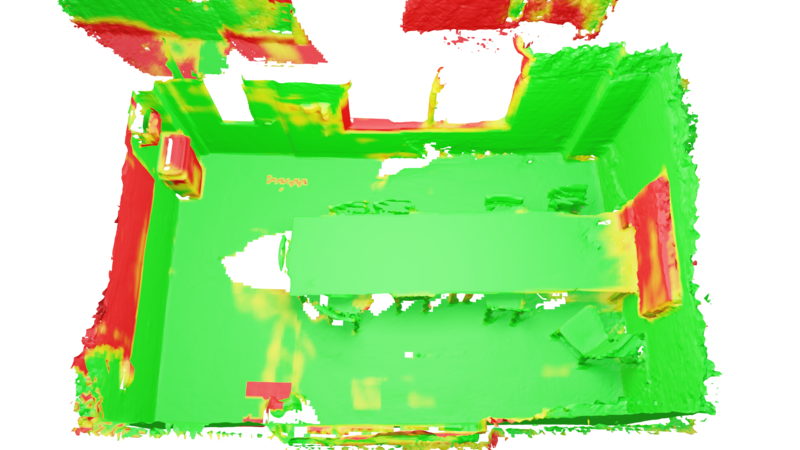}
    \caption{Scene 0599}
\end{subfigure}
\begin{subfigure}[t]{0.15\textwidth}
\vspace{-2mm}
\pgfplotsset{compat=1.15,
    colormap={mycolormap}{
        color(1)=(red) color(3.5)=(yellow) color(6)=(green)
    }
}
\begin{tikzpicture}
\begin{axis}[
hide axis,
    scale only axis,
    height=0pt,
    width=0pt,
    point meta min=1,
    point meta max=6,
    colorbar,
    colorbar style={
        height=2cm,
        width=3mm,
        ylabel={\scriptsize max logit},
        ytick={1, 2, 3, 4, 5, 6},
        yticklabel style={
          font=\scriptsize
        }
    }]
\addplot [draw=none] coordinates {(0,0)};
\end{axis}
\end{tikzpicture}
\end{subfigure}
\vspace{-2mm}
    \caption{Uncertainty in the volumetric maps, measured as the max-logit (lower is more uncertain). As expected, the outlier objects towel (a), books (b), and TV (c) have high uncertainty. Note that some other objects like the ladder in (a) that is also shown in Figure~\ref{fig:more_objects} have high uncertainty.}
    \label{fig:uncertainty_maps}
    \vspace{-5mm}
\end{figure}
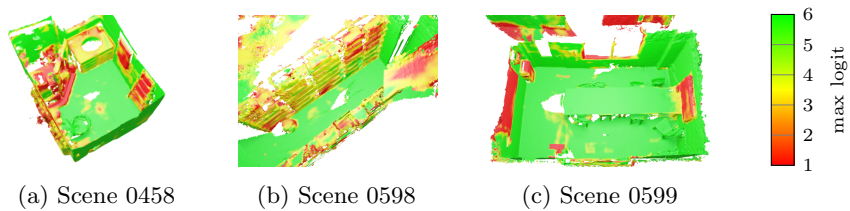

\begin{table}[t]
    \centering
\resizebox{\linewidth}{!}{%
    \begin{tabular}{ll|cccc|cccccccc|cc}
\toprule
 & & \multicolumn{4}{c|}{information} & \multicolumn{8}{c|}{single class IoU}\\
    & variant & 3D & segm & imgn & geom & wall &  floor &  chair &  table &  door &  window & \textbf{tv} &  whiteboard &  mIoU &  v score\\
\midrule

& base model & - & - & - & - & 81 &     72 &     67 &     84 &    30 &      12 &           0 &          30 &    47 &       62 \\
& semantic map &  \faCheck & - & - & - & 82 &     78 &     73 &     89 &    33 &       8 &           0 &          51 &    52 &       68\\
\midrule
\multirow{2}{*}{clustering} & seg only & - & 1 & - & - &  51 &      5 &     11 &      3 &     0 &       0 &          28 &           0 &    12 &       27 \\ % hdbscan3539
%& imgnet only & - & - & 1 & - &     9 &     15 &      6 &     25 &     9 &      11 &          23 &           0 &    12 &       25 &         6 \\ % hdbscanimgnet3546
%& dino only & - & - & - & - &    27 &     22 &     21 &     61 &    25 &       0 &          39 &          31 &    28 &       41 &        20 \\ % hdbscandino3549
%& seg + geom & - & 1 & - & 0 & 54 &     14 &      0 &     10 &     0 &       0 &           0 &           0 &    10 &       29 &        16 \\ % segandgeo3478 
%& imgn + geom & - & - & .86 & .14 &  20 &     28 &     10 &     37 &     0 &       0 &          42 &           0 &    17 &       22 &         7\\ % imgandgeohdbscan3544
& seg + imgn & - & .69 & .31 & - & 32 &     32 &     47 &     27 &     10 &       1 &          27 &           7 &    23 &       40  \\ % segandimgnethdbscan3390
%& seg + geom + imgn & - & .79 & .21 & 0 &    51 &     67 &     42 &     72 &     0 &       0 &          27 &          35 &    37 &       50 &        25 \\ % seggeoimghdbscan3403
%& seg + geom + dino & & & & &   20 &     32 &      0 &     27 &     0 &       0 &           0 &           0 &    10 &       14 &        32 \\ % seggeodinohdbscan3524
\midrule
\multirow{2}{*}{pseudolabel}
& seg only & \faCheck & 1 & - & - & 83 &     78 &     73 &     91 &    37 &      21 &          42 &          20 &    56 &       68 \\ % merged2-pseudolabel-pred-hdbscan3539
%& imgnet only & \faCheck & - & 1 & - &    84 &     78 &     73 &     91 &    37 &      11 &          28 &          17 &    52 &       68 &         6 \\ % merged2-pseudolabel-pred-hdbscanimgnet3546
%& dino only & \faCheck & - & - & - &    84 &     78 &     73 &     91 &    37 &      14 &          48 &          40 &    58 &       69 &        20 \\ % merged2-pseudolabel-pred-hdbscandino3549
%& seg + geom & \faCheck & 1 & - & 0 & 83 &     78 &     73 &     91 &    37 &      13 &          40 &          24 &    55 &       68\\ % segandgeo 3478
% & imgn + geom & \faCheck & - & .86 & .14     &    84 &     78 &     73 &     91 &    37 &      26 &          46 &          16 &    56 &       68 &         7 \\ % merged2-pseudolabel-pred-imgandgeohdbscan3544
 & seg + imgn & \faCheck & .69 & .31 & -   &    84 &     78 &     73 &     91 &    37 &      19 &          43 &          21 &    56 &       69\\ % merged-pseudolabel-pred-segandimgnethdbscan3390
%& seg + geom + imgn & \faCheck & .79 & .21 & 0       &    83 &     78 &     73 &     87 &    37 &       2 &          33 &          29 &    53 &       67 &        25 \\ % merged-pseudolabel-pred-seggeoimghdbscan3403
%& seg + geom + dino & & & & &    84 &     78 &     73 &     91 &    37 &      15 &          64 &          16 &    57 &       69 &        32 \\ % merged2-pseudolabel-pred-seggeodinohdbscan3524
\bottomrule
\end{tabular}
}\vspace{3pt}
    \caption{Ablation of different information sources on scene 0354\_00. Listed in `information' are the weights of different feature distances.}
    \label{tab:ablation354}
    \vspace{-5mm}
\end{table}

\begin{table}[t]
    \centering
\resizebox{\linewidth}{!}{%
    \begin{tabular}{ll|cccc|cccccccc|cc}
\toprule
 & & \multicolumn{4}{c|}{information} & \multicolumn{8}{c|}{single class IoU}\\
    & variant & 3D & segm & imgn & geom & wall &  floor &  cabinet &  door &  mirror &  ceiling &  towel &  sink &  mIoU &  v score\\
\midrule
& base model & - & - & - & - & 53 &     76 &       76 &    18 &      10 &       61 &      0 &    62 &    45 &       56  \\
& semantic map & \faCheck & - & - & - & 50 &     83 &       77 &    11 &      14 &       60 &      0 &    57 &    44 &       59\\
\midrule
\multirow{3}{*}{clustering}
& seg only & - & 1 & - & - & 37 &     33 &        0 &     0 &       0 &        0 &      0 &     0 &     9 &       25\\ %%3319
& seg + geo & - & .94 & - & .06 &    23 &     31 &       71 &    26 &       22 &        19 &     12 &     19 &    28 &       34\\ % segandgeohdbscan3467
& seg + imgn + geo & - & .64 & .15 & .21 &  31 &     22 &        0 &     11 &       0 &        0 &      0 &     0 &     8 &       23 \\ % seggeoimghdbscan3335
\midrule
\multirow{3}{*}{pseudolabel}
& seg only & \faCheck & 1 & - & - & 43 &     80 &       50 &    11 &       0 &       58 &      4 &    84 &    41 &       48\\ % merged2-pseudolabel-pred-hdbscan3319      
& seg + geo & \faCheck & .94 & - & .06 &  48 &     80 &       72 &    11 &       23 &       58 &     16 &    84 &    49 &       52  \\ % merged2-pseudolabel-pred-segandgeohdbscan3467
& seg + imgn + geo & \faCheck& .64 & .15 & .21 &  48 &     80 &       50 &    11 &       5 &       58 &     32 &    84 &    46 &       53 \\ % merged2-pseudolabel-pred-seggeoimghdbscan3335
\bottomrule
\end{tabular}
}\vspace{3pt}
    \caption{Ablation of different information sources on scene 0574\_00. Listed in `information' are the weights of different feature distances.}
    \label{tab:ablation574}
\vspace{-7mm}
\end{table}

\subsection{Design Choice Verification}
We now evaluate different design choices. In Figure~\ref{fig:uncertainty_maps}, we show the volumetric maps of different environments and the mapped uncertainty estimation. These maps show that the uncertainty in combination with mapping is very effective to identify uncertain parts of the scene. Next to the `target' objects, also other uncertain objects are identified. It is expected that a model can also be uncertain about known classes, or that more than 1 novel object are present is a scene. We further show this in Figure~\ref{fig:more_objects}.

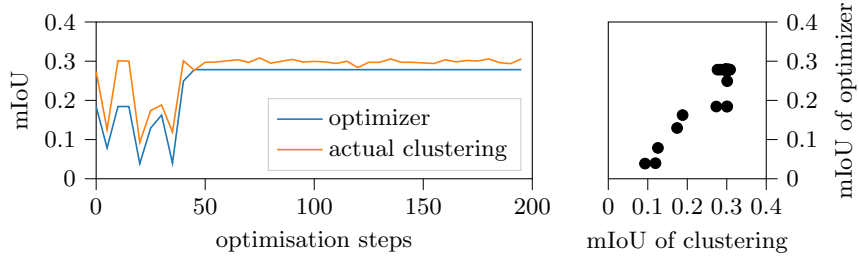
\begin{figure}[t]
\vspace{-2mm}
    \centering
    % This file was created with tikzplotlib v0.10.1.
\begin{tikzpicture}
\footnotesize

\definecolor{darkgray176}{RGB}{176,176,176}
\definecolor{darkorange25512714}{RGB}{255,127,14}
\definecolor{lightgray204}{RGB}{204,204,204}
\definecolor{steelblue31119180}{RGB}{31,119,180}

\begin{groupplot}[group style={group size=2 by 1}]
\nextgroupplot[
height=0.3\linewidth,
width=0.6\linewidth,
legend cell align={left},
legend style={fill opacity=0.8, draw opacity=1, text opacity=1, draw=lightgray204, at={(.975,.05)},anchor=south east},
tick align=outside,
tick pos=left,
x grid style={darkgray176},
xmin=0, xmax=200,
xtick style={color=black},
y grid style={darkgray176},
ymin=0, ymax=0.4,
ytick style={color=black},
xlabel={\footnotesize optimisation steps},
ylabel={\footnotesize mIoU}
]
\addplot [semithick, steelblue31119180]
table {%
0 0.184313239674431
5 0.0786495559887373
10 0.184313239674431
15 0.184313239674431
20 0.0386935450608373
25 0.129562344114748
30 0.162479502091254
35 0.0397331454311288
40 0.249042592681856
45 0.278408958209764
50 0.278408958209764
55 0.278408958209764
60 0.278408958209764
65 0.278408958209764
70 0.278408958209764
75 0.278408958209764
80 0.278408958209764
85 0.278408958209764
90 0.278408958209764
95 0.278408958209764
100 0.278408958209764
105 0.278408958209764
110 0.278408958209764
115 0.278408958209764
120 0.278408958209764
125 0.278408958209764
130 0.278408958209764
135 0.278408958209764
140 0.278408958209764
145 0.278408958209764
150 0.278408958209764
155 0.278408958209764
160 0.278408958209764
165 0.278408958209764
170 0.278408958209764
175 0.278408958209764
180 0.278408958209764
185 0.278408958209764
190 0.278408958209764
195 0.278408958209764
};
\addlegendentry{optimizer}
\addplot [semithick, darkorange25512714]
table {%
0 0.273542928454921
5 0.12572086740327
10 0.301096127520941
15 0.300240780935036
20 0.0929470387859802
25 0.174221964275598
30 0.188154406344841
35 0.119430655265058
40 0.301144439485482
45 0.276718127839856
50 0.297150976848981
55 0.297812082624987
60 0.300989613544365
65 0.303658114742913
70 0.296991145858942
75 0.308319900115153
80 0.295053206932063
85 0.299876382264651
90 0.304536324594205
95 0.297812082624987
100 0.299481602037414
105 0.29824183815039
110 0.294151455094302
115 0.299876382264651
120 0.283817030726992
125 0.297150976848981
130 0.297150976848981
135 0.305837357583943
140 0.297150976848981
145 0.297150976848981
150 0.295592367638706
155 0.294092488652164
160 0.303658114742913
165 0.298207943585144
170 0.302043009118569
175 0.300534340850266
180 0.305837357583943
185 0.296339891139384
190 0.293702470833026
195 0.305837357583943
};
\addlegendentry{actual clustering}

\nextgroupplot[
height=0.3\linewidth,
width=0.3\linewidth,
%scaled y ticks=manual:{}{\pgfmathparse{#1}},
tick align=outside,
ytick pos=right,
xtick pos=left,
x grid style={darkgray176},
xmin=0, xmax=0.4,
xtick style={color=black},
y grid style={darkgray176},
ymin=0, ymax=0.4,
ytick style={color=black},
xlabel={\footnotesize mIoU of clustering},
ylabel={\footnotesize mIoU of optimizer}
]
\addplot [draw=black, fill=black, mark=*, only marks]
table{%
x  y
0.273542928454921 0.184313239674431
0.12572086740327 0.0786495559887373
0.301096127520941 0.184313239674431
0.300240780935036 0.184313239674431
0.0929470387859802 0.0386935450608373
0.174221964275598 0.129562344114748
0.188154406344841 0.162479502091254
0.119430655265058 0.0397331454311288
0.301144439485482 0.249042592681856
0.276718127839856 0.278408958209764
0.297150976848981 0.278408958209764
0.297812082624987 0.278408958209764
0.300989613544365 0.278408958209764
0.303658114742913 0.278408958209764
0.296991145858942 0.278408958209764
0.308319900115153 0.278408958209764
0.295053206932063 0.278408958209764
0.299876382264651 0.278408958209764
0.304536324594205 0.278408958209764
0.297812082624987 0.278408958209764
0.299481602037414 0.278408958209764
0.29824183815039 0.278408958209764
0.294151455094302 0.278408958209764
0.299876382264651 0.278408958209764
0.283817030726992 0.278408958209764
0.297150976848981 0.278408958209764
0.297150976848981 0.278408958209764
0.305837357583943 0.278408958209764
0.297150976848981 0.278408958209764
0.297150976848981 0.278408958209764
0.295592367638706 0.278408958209764
0.294092488652164 0.278408958209764
0.303658114742913 0.278408958209764
0.298207943585144 0.278408958209764
0.302043009118569 0.278408958209764
0.300534340850266 0.278408958209764
0.305837357583943 0.278408958209764
0.296339891139384 0.278408958209764
0.293702470833026 0.278408958209764
0.305837357583943 0.278408958209764
};
\end{groupplot}

\end{tikzpicture}
    \vspace{-3mm}
    \caption{Analysis of the correlation between the optimisation objective (clustering subset measured against high-confidence predictions) and the actual performance of the clustering (full images measured against ground-truth labels).}
    \label{fig:optimisation}
    \vspace{-5mm}
\end{figure}

As described in Section~\ref{sub:clustering}, we choose clustering parameters based on an optimisation objective.  Figure~\ref{fig:optimisation} shows an analysis of whether this objective, which only approximates the true clustering performance, correlates with the true performance of the clustering. For this analysis, we subsample every 20th frame and cluster segmentation features only, to then evaluate the full clustering performance at every 5th optimisation step. We observe that the correlation increases in noise for higher mIoU, but in general correlates well enough to disregard bad parameters.

Tables~\ref{tab:ablation354} and~\ref{tab:ablation574} investigate how beneficial multiple sources of information are in the clustering problem. Firstly, the tables show that clustering with the segmentation features does not result in the same performance as the prediction of the same network, indicating that significant knowledge about classes is stored in the final layer (we extract features before the final layer). Over both scenes, we see that multiple sources of information result in better pseudolabels than just the segmentation features. We also see that which information is useful differs from scene to scene, motivating our decision to select the weights of the sources in each scene through optimisation. While the combination of all features therefore creates a robust clustering for different scenes, Table~\ref{tab:ablation574} shows that this can result in a small tradeoff in single scene performance. We assume that this is caused by the increase of the optimisation space for every new feature.

\section{Discussion \& Outlook}
We investigate the problem of semantic scene understanding in unknown environments containing novel objects. We develop a framework of clustering, inference, and mapping that can be used to autonomously discover novel categories and improve semantic knowledge. It generalises over existing work and helps us to create a new method based on black-box optimisation and information fusion. 

To discover novel categories, our experiments show that unsupervised features are as useful as supervised ones, especially when multiple features are used. We also show that prior knowledge is very helpful, e.g. to optimize parameters. Better (gradient based) optimisation and a less noisy objective function may even improve this mechanism. In general, from the components in Figure~\ref{fig:systemdiagram}, most are able to propagate gradients, opening opportunities for future research to replace more heuristics with deep learning.

While we showed that information fusion can be advantageous, it requires graph clustering, which is neither mini-batch compatible nor differentiable. We do not report runtimes, because all steps after mapping are not required to be online and our implementations are not optimised for this. We can however report that the meta segmentation of~\cite{uhlemeyer_towards_2022} and our optimisation usually required multiple hours. Both points show the potential of more efficient clustering.

This work did not touch upon the questions of continual learning or active exploration. The first asks how to organise class-incremental learning such that more and more semantic categories can be discovered as a robot moves from scene to scene. Prior work as touched on this topic~\cite{blum_self-improving_2021,michieli_incremental_2019,uhlemeyer_towards_2022}, but none has evaluated self-supervised approaches with multiple classes over multiple consecutive environments. Similarly, the influence of actively planning trajectories to aid discovery remains to be investigated, with promising results from domain adaptation on known classes indicating that planning is helpful~\cite{zurbrugg_embodied_2022,chaplot_seal_2021}.

Our experiments had to overcome a lack of available implementations and quality problems in the data. By releasing our segmentation models, implementations of related work, and implementation of our SCIM approach, we aim to accelerate future research on this topic.
% ---- Bibliography ----
%

\printbibliography
%\bibliographystyle{IEEEtran}
%\bibliography{references}

\clearpage
\appendix

\section{Qualitative Examples}
\begin{figure}
    \centering
\resizebox{\linewidth}{!}{%
\begin{tikzpicture}[
every node/.style={outer sep=0pt, inner sep=0pt},
node distance=0mm,
]
\node[label={base model\strut}] (354-1) {\includegraphics[width=3cm]{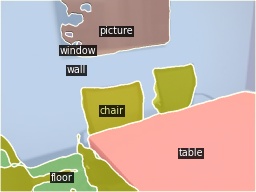}};
\node[left=of 354-1] {\rotatebox{90}{0354\strut}};
\node[label={adapted nakajima\strut}, right=of 354-1] (354-2) {\includegraphics[width=3cm]{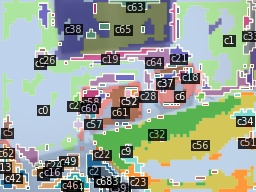}};
\node[label={adapted uhlemeyer\strut}, right=of 354-2] (354-3) {\includegraphics[width=3cm]{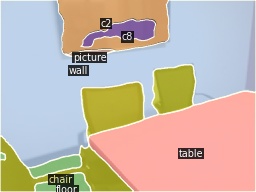}};
\node[label={SCIM with imgnet\strut}, right=of 354-3] (354-4) {\includegraphics[width=3cm]{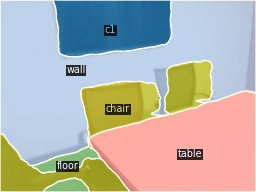}};
\node[label={SCIM unsupervised\strut}, right=of 354-4] {\includegraphics[width=3cm]{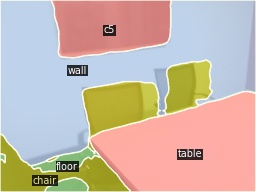}};

\node[below=of 354-1] (575-1) {\includegraphics[width=3cm]{figures/frames/scene0575_00_002521_pred.jpg}};
\node[left=of 575-1] {\rotatebox{90}{0575\strut}};
\node[right=of 575-1] (575-2) {\includegraphics[width=3cm]{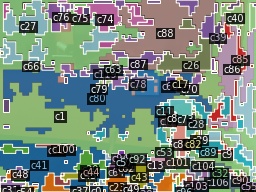}};
\node[right=of 575-2] (575-3) {\includegraphics[width=3cm]{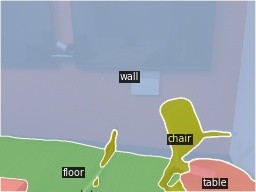}};
\node[right=of 575-3] (575-4) {\includegraphics[width=3cm]{figures/frames/scene0575_00_002521_pred3287-merged-pseudolabel-pred-segandimgnethdbscan3060.jpg}};
\node[right=of 575-4] (575-5) {\includegraphics[width=3cm]{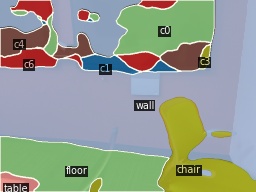}};

\node[below=of 575-1] (599-1) {\includegraphics[width=3cm]{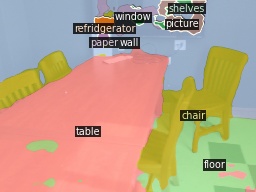}};
\node[left=of 599-1] {\rotatebox{90}{0599\strut}};
\node[right=of 599-1] (599-2) {\includegraphics[width=3cm]{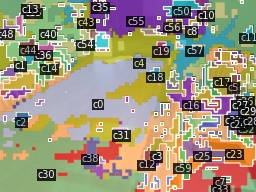}};
\node[right=of 599-2] (599-3) {\includegraphics[width=3cm]{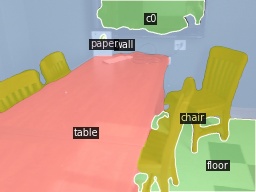}};
\node[right=of 599-3] (599-4) {\includegraphics[width=3cm]{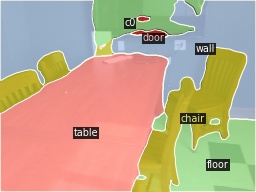}};
\node[right=of 599-4] {\includegraphics[width=3cm]{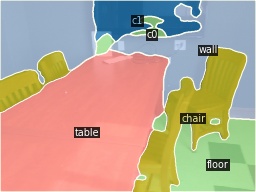}};

\node[below=of 599-1] (598-1) {\includegraphics[width=3cm]{figures/frames/scene0598_02_000381_pred.jpg}};
\node[left=of 598-1] {\rotatebox{90}{0598\strut}};
\node[right=of 598-1] (598-2) {\includegraphics[width=3cm]{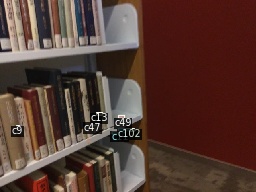}};
\node[right=of 598-2] (598-3) {\includegraphics[width=3cm]{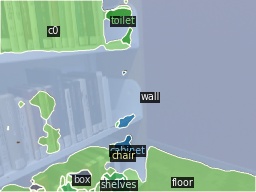}};
\node[right=of 598-3] (598-4) {\includegraphics[width=3cm]{figures/frames/scene0598_02_000381_pred3404-merged-pseudolabel-pred-seggeoimghdbscan3274.jpg}};
\node[right=of 598-4] {\includegraphics[width=3cm]{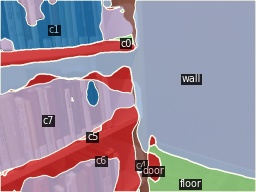}};

\node[below=of 598-1] (458-1) {\includegraphics[width=3cm]{figures/frames/scene0458_00_001841_pred.jpg}};
\node[left=of 458-1] {\rotatebox{90}{0458\strut}};
\node[right=of 458-1] (458-2) {\includegraphics[width=3cm]{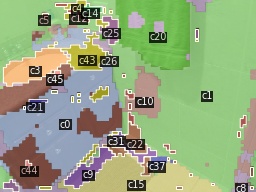}};
\node[right=of 458-2] (458-3) {\includegraphics[width=3cm]{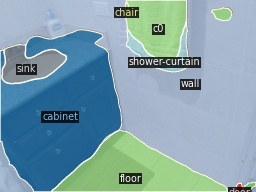}};
\node[right=of 458-3] (458-4) {\includegraphics[width=3cm]{figures/frames/scene0458_00_001841_pred3435-merged-pseudolabel-pred-segandimgnethdbscan3298.jpg}};
\node[right=of 458-4] {\includegraphics[width=3cm]{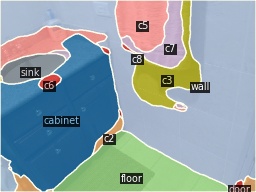}};

\node[below=of 458-1] (574-1) {\includegraphics[width=3cm]{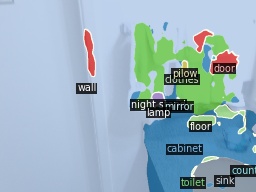}};
\node[left=of 574-1] {\rotatebox{90}{0574\strut}};
\node[right=of 574-1] (574-2) {\includegraphics[width=3cm]{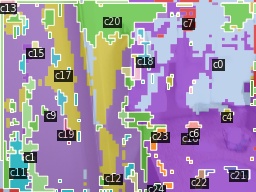}};
\node[right=of 574-2] (574-3) {\includegraphics[width=3cm]{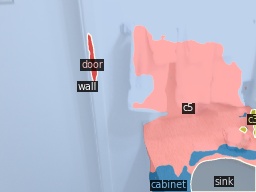}};
\node[right=of 574-3] (574-4) {\includegraphics[width=3cm]{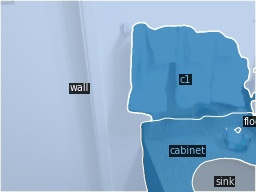}};
\node[right=of 574-4] {\includegraphics[width=3cm]{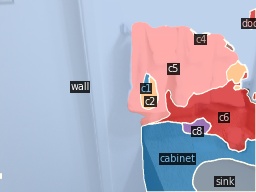}};
\end{tikzpicture}}
    \caption{Qualitative examples corresponding to the results in Table~\ref{tab:results}.}
    \label{fig:qualitative1}
\end{figure}

\begin{figure}
    \centering
\resizebox{\linewidth}{!}{%
\begin{tikzpicture}[
every node/.style={outer sep=0pt, inner sep=0pt},
node distance=0mm,
]
\node[label={base model\strut}] (25-1) {\includegraphics[width=3cm]{figures/frames/scene0354_00_000621_pred.jpg}};
\node[left=of 25-1] {\rotatebox{90}{0025\strut}};
\node[label={adapted nakajima\strut}, right=of 25-1] (25-2) {\includegraphics[width=3cm]{figures/frames/scene0354_00_000621_nakajima3440.jpg}};
\node[label={adapted uhlemeyer\strut}, right=of 25-2] (25-3) {\includegraphics[width=3cm]{figures/frames/scene0354_00_000621_pred3472-uhlemeyer3457.jpg}};
\node[label={SCIM with imgnet\strut}, right=of 25-3] (25-4) {\includegraphics[width=3cm]{figures/frames/scene0354_00_000621_pred3417-merged-pseudolabel-pred-seggeoimghdbscan3403.jpg}};
\node[label={SCIM unsupervised\strut}, right=of 25-4] {\includegraphics[width=3cm]{figures/frames/scene0354_00_000621_pred3552-merged2-pseudolabel-pred-seggeodinohdbscan3524.jpg}};

\node[below=of 25-1] (164-1) {\includegraphics[width=3cm]{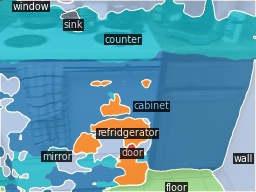}};
\node[left=of 164-1] {\rotatebox{90}{0164\strut}};
%\node[right=of 164-1] (164-2) {\includegraphics[width=3cm]{figures/frames/164.jpg}};
%\node[right=of 164-2] 
\node[below=of 25-3]
(164-3) {\includegraphics[width=3cm]{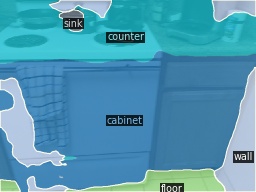}};
\node[right=of 164-3] (164-4) {\includegraphics[width=3cm]{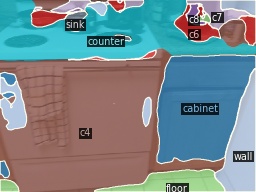}};
%\node[right=of 164-4] (164-5) {\includegraphics[width=3cm]{figures/frames/164-merged2-pseudolabel-pred-seggeodinohdbscan3526.jpg}};

\node[below=of 164-1] (568-1) {\includegraphics[width=3cm]{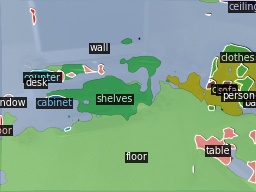}};
\node[left=of 568-1] {\rotatebox{90}{0568\strut}};
\node[right=of 568-1] (568-2) {\includegraphics[width=3cm]{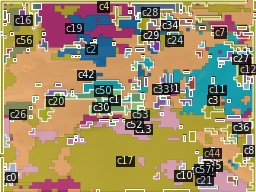}};
\node[right=of 568-2] (568-3) {\includegraphics[width=3cm]{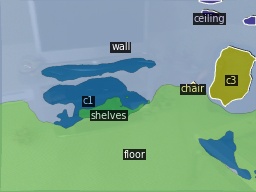}};
\node[right=of 568-3] (568-4) {\includegraphics[width=3cm]{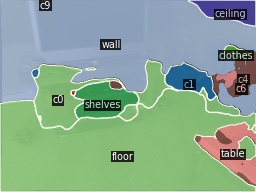}};
\node[right=of 568-4] (568-5) {\includegraphics[width=3cm]{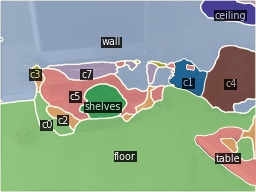}};
\end{tikzpicture}}
    \caption{Qualitative examples corresponding to the results in Table~\ref{tab:unsuccessful_scenes}.}
    \label{fig:qualitative2}
\end{figure}

We show qualitative examples in Figures~\ref{fig:qualitative1} and~\ref{fig:qualitative2}.

\section{Selection of Evaluation Scenes}
We download the first 100 validation scenes of ScanNet and find in there 6 scenes with television, 5 scenes with towel and 4 scenes with books. To account for mistakes in the labelling and other data quality issues, we further had to filer them:
\begin{itemize}
    \item We remove scenes 0426 and 0608 (both with tvs) because of too many novel objects without annotation and severe degradation of the base model performance (see Table~\ref{tab:unsuccessful_scenes}).
    \item Scenes 0665 and 0565 have annotations for a book, but the objects are actually a cable outlet (0665) and a carton box (0565).
    \item Scene 0025 has two books, but they are very small and no method is able to pick it up (see Table~\ref{tab:unsuccessful_scenes}). 
    \item The towel in scene 0164 is visually fully enclosed by another unknown category (stove) without annotation, making it impossible to measure whether the towel and stove are separated in the clustering. We show some results in Table~\ref{tab:unsuccessful_scenes}.
    \item We ignore the label `bookshelf' in scene 0598 because some books are annotated as `books' and others as `bookshelf'. We do not want to punish methods for classifying all books in the same cluster.
\end{itemize}

\begin{table}[t]
    \centering
    \begin{tabular}{lllrr}
\toprule
 & & &  \multicolumn{2}{c}{training traj.}\\ 
outlier & scene & method & out IoU & mIoU\\
 \midrule
\multirow{1}{*}{tv} & \multirow{1}{*}{0426} & 
base model & 0 & 36\\
\midrule
\multirow{6}{*}{tv} & \multirow{6}{*}{0568} & 
base model & 0 & 50\\
& & adapted nakajima, segm. only & 9 & 31\\ % 3567
& & SCIM fusing segm. + geom. + dino & 0 & 52\\ % 3530
& & adapted uhlemeyer & 0 & 53 \\ % 3423
& & adapted uhlemeyer + map & 0 & 60 \\ % 2425
& & SCIM fusing segm. + geom. + imgn. & 0 & 54\\ % 3424
\midrule
\multirow{1}{*}{tv} & \multirow{1}{*}{0608} & 
base model & 0 & 45\\
 \midrule
\multirow{1}{*}{books} & \multirow{1}{*}{0025} & 
base model & 0 & 42\\
& & adapted uhlemeyer & 0 & 44\\ % 3251
& & SCIM fusing segm. + geom. + imgn.  & 8 & 49\\
\midrule
\multirow{5}{*}{towel} & \multirow{5}{*}{0164} & 
base model & 0 & 50\\
& & adapted nakajima, segm. only & 6 & 25\\ %3566
& & adapted uhlemeyer & 12 & 48\\ % 3385
& & adapted uhlemeyer + map & 0 & 53\\
& & SCIM fusing segm. + geom. + imgn.  & 10 & 54\\ % 3398
\bottomrule
    \end{tabular}
    \caption{Challenging Scenes in which no method achieves good detection of the outlier, often due to clutter.}
    \label{tab:unsuccessful_scenes}
\end{table}
\end{document}